\def\year{2021}\relax
\pdfoutput=1
\documentclass[letterpaper]{article} 
\usepackage{aaai21}  
\usepackage{times}  
\usepackage{helvet} 
\usepackage{courier}  
\usepackage[hyphens]{url}  
\usepackage{graphicx} 
\usepackage{natbib}  
\usepackage{caption} 
\usepackage{microtype}
\usepackage{latexsym}
\usepackage{tikz}
\usepackage{tikz-qtree}
\usepackage{pgfplots}
\usepackage{subfigure}
\usepackage{xcolor}
\usepackage{array, multirow} 
\usepackage{amssymb}
\usepackage{booktabs}
\usepackage{verbatim}

\usetikzlibrary{backgrounds,fit}
\usetikzlibrary{shapes,arrows,shadows}
\usetikzlibrary{patterns}
\usetikzlibrary{arrows,decorations.pathreplacing}
\usetikzlibrary{shadows} 

\usepgflibrary{arrows} 
\usetikzlibrary{arrows} 
\usetikzlibrary{decorations}
\usetikzlibrary{arrows,shapes}

\usetikzlibrary{positioning,fit,calc}

\usetikzlibrary{mindmap,backgrounds} 

\usetikzlibrary{arrows,decorations}
\usetikzlibrary{decorations.pathreplacing}
\usetikzlibrary{positioning,calc}
\usetikzlibrary{datavisualization}

\usetikzlibrary{pgfplots.groupplots}
\usetikzlibrary{patterns}

\newlength{\wseg}
\newlength{\hseg}
\newlength{\wnode}
\newlength{\hnode}
\newlength{\base}

\definecolor{ugreen}{rgb}{0,0.5,0}
\definecolor{ublue}{RGB}{0,31,239}
\definecolor{myblue}{RGB}{97,214,255}

\title{Learning Light-Weight Translation Models from Deep Transformer}
\author{Bei Li\textsuperscript{\rm 1}\thanks{Equal contribution},
  Ziyang Wang\textsuperscript{\rm 1}$^{*}$,
  Hui Liu\textsuperscript{\rm 1}$^{*}$,
	Quan Du\textsuperscript{\rm 1,2},\\
  Tong Xiao\textsuperscript{\rm 1,2}\thanks{Corresponding author},
  Chunliang Zhang\textsuperscript{\rm 1,2},
  Jingbo Zhu\textsuperscript{\rm 1,2}\\
}
\affiliations{

    \textsuperscript{\rm 1}NLP Lab, School of Computer Science and Engineering,
    Northeastern University, Shenyang, China\\
    \textsuperscript{\rm 2}NiuTrans Research, Shenyang, China\\

    \{libei\_neu,duquanneu\}@outlook.com,
    \{wangziyang,huiliu\}@stumail.neu.edu.cn,
    \{xiaotong,zhangcl,zhujingbo\}@mail.neu.edu.cn

}

\begin{document}

\maketitle

\begin{abstract}
    Recently, deep models have shown tremendous improvements in neural machine translation (NMT). However, systems of this kind are computationally expensive and memory intensive. In this paper, we take a natural step towards learning strong but light-weight NMT systems. We proposed a novel group-permutation based knowledge distillation approach to compressing the deep Transformer model into a shallow model. The experimental results on several benchmarks validate the effectiveness of our method. Our compressed model is $8\times$ shallower than the deep model, with almost no loss in BLEU. To further enhance the teacher model, we present a \textit{Skipping Sub-Layer} method to randomly omit sub-layers to introduce perturbation into training, which achieves a BLEU score of 30.63 on English-German \textit{newstest2014}. The code is publicly available at https://github.com/libeineu/GPKD.
\end{abstract}

\section{Introduction}

Neural machine translation (NMT) has advanced significantly in recent years \cite{bahdanau2014neural}. In particular, the Transformer model has become popular for its well-designed architecture and the ability to capture the dependency among positions over the entire sequence \cite{vaswani2017attention}. Early systems of this kind stack 4-8 layers on both the encoder and decoder sides \cite{wu2016google,gehring2017convs2s}, and the improvement often comes from the use of wider networks (a.k.a., Transformer-Big). More recently, researchers try to explore deeper models for Transformer. Encouraging results appeared in architecture improvements by creating direct pass from the low-level encoder layers to the decoder \cite{bapna-etal-2018-training,wang-etal-2019-learning,wei2004multiscale,wu-etal-2019-depth, li-etal-2019-niutrans}, and proper initialization strategies \cite{zhang-etal-2019-improving,xu2019lipschitz,liu2020understanding,Huang2020improving}.

Despite promising improvements, problems still remain in deep NMT. Deep Transformer stacked by dozens of encoder layers always have a large number of parameters, which are computationally expensive and memory intensive. For example, a 48-layer Transformer is $3\times$ larger than a 6-layer system and $1.5\times$ slower for inference. It is difficult to deploy such models on resource-restricted devices, such as mobile phones. Therefore, it is crucial to compress such heavy systems into light-weight ones while keeping their performance.

Knowledge distillation is a promising method to address the issue. Although several studies \cite{sun-etal-2019-patient, jiao2019tinybert} have attempted to compress the 12-layer BERT model through knowledge distillation, effectively compressing extremely deep Transformer NMT systems is still an open question in the MT community. In addition, these methods leverage sophisticated layer-wise distillation loss functions to minimize the distance between the teacher and the student models, which requires huge memory consumption and enormous training cost.

In this paper, we investigate simple and efficient compression strategies for deep Transformer. We propose a novel Transformer compression approach (named as group-permutation based knowledge distillation method (\textsc{Gpkd})) to transfer the knowledge from an extremely deep teacher model into a shallower student model. We disturb the computation order among each layer group during the teacher training phase, which is easy to implement and memory friendly. Moreover, to further enhance the performance of the teacher network, we introduce a vertical ``dropout'' (named as skipping sub-layer method) into training by randomly omitting sub-layers to prevent co-adaptations of the over-parameterized teacher network. Although similar technique has been discussed in \citet{fan2019reducing}'s work, we believe that the finding here is complementary to theirs. Both \textsc{Gpkd} and regularization training methods can be well incorporated into the teacher training process, which is essential for obtaining a strong but light-weight student model.

\pgfdeclarepatternformonly{soft horizontal lines}{\pgfpointorigin}{\pgfqpoint{100pt}{1pt}}{\pgfqpoint{100pt}{3pt}}%
{
  \pgfsetstrokeopacity{0.3}
  \pgfsetlinewidth{0.1pt}
  \pgfpathmoveto{\pgfqpoint{0pt}{0.5pt}}
  \pgfpathlineto{\pgfqpoint{100pt}{0.5pt}}
  \pgfusepath{stroke}
}

\pgfdeclarepatternformonly{soft crosshatch}{\pgfqpoint{-1pt}{-1pt}}{\pgfqpoint{6pt}{6pt}}{\pgfqpoint{5pt}{5pt}}%
{
  \pgfsetstrokeopacity{0.3}
  \pgfsetlinewidth{0.4pt}
  \pgfpathmoveto{\pgfqpoint{4.5pt}{0pt}}
  \pgfpathlineto{\pgfqpoint{0pt}{4.5pt}}
  \pgfpathmoveto{\pgfqpoint{0pt}{0pt}}
  \pgfpathlineto{\pgfqpoint{4.5pt}{4.5pt}}
  \pgfusepath{stroke}
}

\definecolor{ugreen}{rgb}{0,0.5,0}

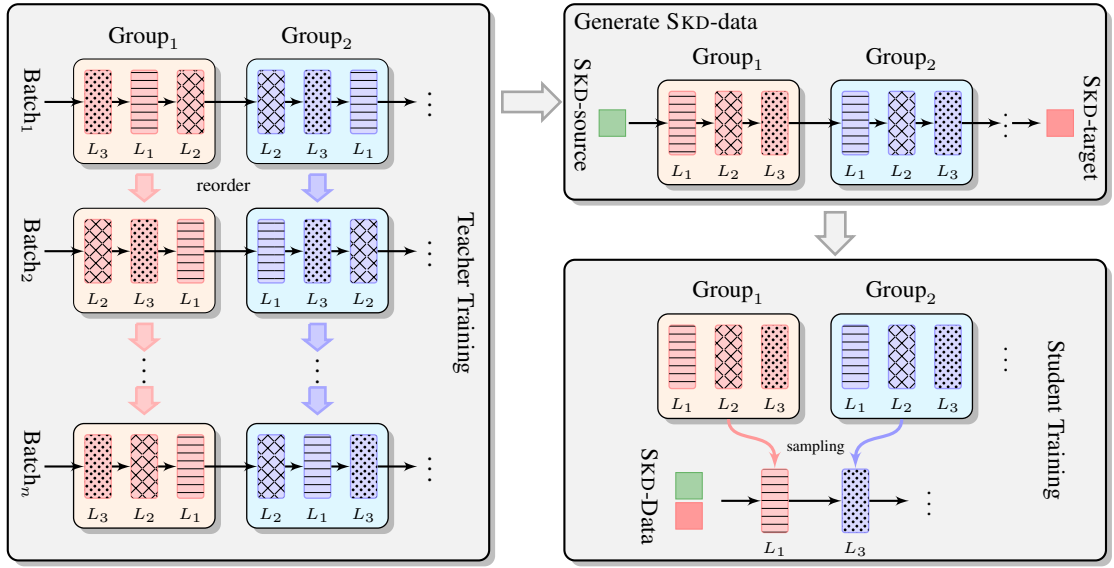
\begin{figure*}
  \centering
  \begin{tikzpicture}
  \setlength{\base}{0.7em};

  \tikzstyle{layer_node} = [line width=0.5pt,minimum height=2.4em,minimum width = 1em,inner sep=3pt,rounded corners=1pt,font=\tiny]
  \tikzstyle{legend_node} = [minimum height=1em,minimum width = 1em,draw,thick,inner sep=3pt,rounded corners=1pt,font=\footnotesize]
  \node[layer_node,draw=red!40,fill=red!20] (layer_1) at (0,0) {};
  \node[layer_node,pattern = crosshatch dots] at (0,0) {};
  \node[font=\tiny,anchor = north] (layer_1_id) at (layer_1.south) {$L_{3}$};
  \node[layer_node,anchor=west,draw=red!40,fill=red!20] (layer_2) at ([xshift=\base]layer_1.east) {};
  \node[font=\tiny,anchor = north] (layer_2_id) at (layer_2.south) {$L_{1}$};
  \node[layer_node,anchor=west,pattern = soft horizontal lines] at ([xshift=\base]layer_1.east) {};
  \node[layer_node,anchor = west,draw=red!40,fill=red!20] (layer_3) at ([xshift=\base]layer_2.east) {};
  \node[font=\tiny,anchor = north] (layer_3_id) at (layer_3.south) {$L_{2}$};
  \node[layer_node,anchor=west,pattern = soft crosshatch] at ([xshift=\base]layer_2.east) {};

  \node[layer_node,anchor=west,draw=blue!40,fill=blue!15] (layer_4) at ([xshift=2em]layer_3.east) {};
  \node[font=\tiny,anchor = north] (layer_4_id) at (layer_4.south) {$L_{2}$};
  \node[layer_node,anchor=west,pattern = soft crosshatch] at ([xshift=2em]layer_3.east) {};
  \node[layer_node,anchor=west,draw=blue!40,fill=blue!15] (layer_5) at ([xshift=\base]layer_4.east) {};
  \node[font=\tiny,anchor = north] (layer_5_id) at (layer_5.south) {$L_{3}$};
  \node[layer_node,anchor=west,pattern = crosshatch dots] at ([xshift=\base]layer_4.east) {};
  \node[layer_node,anchor=west,draw=blue!40,fill=blue!15] (layer_6) at ([xshift=\base]layer_5.east) {};
  \node[font=\tiny,anchor = north] (layer_6_id) at (layer_6.south) {$L_{1}$};
  \node[layer_node,anchor=west,pattern = soft horizontal lines] at ([xshift=\base]layer_5.east) {};


  \node[layer_node,anchor=north,draw=red!40,fill=red!20] (layer_2_1) at ([yshift = -3.2em]layer_1.south) {};
  \node[font=\tiny,anchor = north] (layer_2_1_id) at (layer_2_1.south) {$L_{2}$};
  \node[layer_node,anchor=north,pattern = soft crosshatch] at ([yshift = -3.2em]layer_1.south) {};
  \node[layer_node,anchor=west,draw=red!40,fill=red!20] (layer_2_2) at ([xshift=\base]layer_2_1.east) {};
  \node[font=\tiny,anchor = north] (layer_2_2_id) at (layer_2_2.south) {$L_{3}$};
  \node[layer_node,anchor=west,pattern = crosshatch dots] at ([xshift=\base]layer_2_1.east) {};
  \node[layer_node,anchor=west,draw=red!40,fill=red!20] (layer_2_3) at ([xshift=\base]layer_2_2.east) {};
  \node[font=\tiny,anchor = north] (layer_2_3_id) at (layer_2_3.south) {$L_{1}$};
  \node[layer_node,anchor=west,pattern = soft horizontal lines] at ([xshift=\base]layer_2_2.east) {};

  \node[layer_node,anchor=west,draw=blue!40,fill=blue!15] (layer_2_4) at ([xshift=2em]layer_2_3.east) {};
  \node[font=\tiny,anchor = north] (layer_2_4_id) at (layer_2_4.south) {$L_{1}$};
  \node[layer_node,anchor=west,pattern = soft horizontal lines] at ([xshift=2em]layer_2_3.east) {};
  \node[layer_node,anchor=west,draw=blue!40,fill=blue!15] (layer_2_5) at ([xshift=\base]layer_2_4.east) {};
  \node[font=\tiny,anchor = north] (layer_2_5_id) at (layer_2_5.south) {$L_{3}$};
  \node[layer_node,anchor=west,pattern = crosshatch dots] at ([xshift=\base]layer_2_4.east) {};
  \node[layer_node,anchor=west,draw=blue!40,fill=blue!15] (layer_2_6) at ([xshift=\base]layer_2_5.east) {};
  \node[font=\tiny,anchor = north] (layer_2_6_id) at (layer_2_6.south) {$L_{2}$};
  \node[layer_node,anchor=west,pattern = soft crosshatch] at ([xshift=\base]layer_2_5.east) {};

  \node[layer_node,anchor=north,draw=red!40,fill=red!20] (layer_n_1) at ([yshift = -5.7em]layer_2_1.south) {};
  \node[font=\tiny,anchor = north] (layer_n_1_id) at (layer_n_1.south) {$L_{3}$};
  \node[layer_node,anchor=north,pattern = crosshatch dots] at ([yshift = -5.7em]layer_2_1.south) {};
  \node[layer_node,anchor=west,draw=red!40,fill=red!20] (layer_n_2) at ([xshift=\base]layer_n_1.east) {};
  \node[font=\tiny,anchor = north] (layer_n_2_id) at (layer_n_2.south) {$L_{2}$};
  \node[layer_node,anchor=west,pattern = soft crosshatch] at ([xshift=\base]layer_n_1.east) {};
  \node[layer_node,anchor=west,draw=red!40,fill=red!20] (layer_n_3) at ([xshift=\base]layer_n_2.east) {};
  \node[font=\tiny,anchor = north] (layer_n_3_id) at (layer_n_3.south) {$L_{1}$};
  \node[layer_node,anchor=west,pattern = soft horizontal lines] at ([xshift=\base]layer_n_2.east) {};

  \node[layer_node,anchor=west,draw=blue!40,fill=blue!15] (layer_n_4) at ([xshift=2em]layer_n_3.east) {};
  \node[font=\tiny,anchor = north] (layer_n_4_id) at (layer_n_4.south) {$L_{2}$};
  \node[layer_node,anchor=west,pattern = soft crosshatch] at ([xshift=2em]layer_n_3.east) {};
  \node[layer_node,anchor=west,draw=blue!40,fill=blue!15] (layer_n_5) at ([xshift=\base]layer_n_4.east) {};
  \node[font=\tiny,anchor = north] (layer_n_5_id) at (layer_n_5.south) {$L_{1}$};
  \node[layer_node,anchor=west,pattern = soft horizontal lines] at ([xshift=\base]layer_n_4.east) {};
  \node[layer_node,anchor=west,draw=blue!40,fill=blue!15] (layer_n_6) at ([xshift=\base]layer_n_5.east) {};
  \node[font=\tiny,anchor = north] (layer_n_6_id) at (layer_n_6.south) {$L_{3}$};
  \node[layer_node,anchor=west,pattern = crosshatch dots] at ([xshift=\base]layer_n_5.east) {};

  \node[layer_node,anchor=west,draw=red!40,fill=red!20] (layer_t_1) at ([xshift=11em,yshift=-0.8em]layer_6.east) {};
  \node[font=\tiny,anchor = north] (layer_t_1_id) at (layer_t_1.south) {$L_{1}$};
  \node[layer_node,anchor=west,pattern = soft horizontal lines] at ([xshift=11em,yshift=-0.8em]layer_6.east) {};
  \node[layer_node,anchor=west,draw=red!40,fill=red!20] (layer_t_2) at ([xshift=\base]layer_t_1.east) {};
  \node[font=\tiny,anchor = north] (layer_t_2_id) at (layer_t_2.south) {$L_{2}$};
  \node[layer_node,anchor=west,pattern = soft crosshatch] at ([xshift=\base]layer_t_1.east) {};
  \node[layer_node,anchor=west,draw=red!40,fill=red!20] (layer_t_3) at ([xshift=\base]layer_t_2.east) {};
  \node[font=\tiny,anchor = north] (layer_t_3_id) at (layer_t_3.south) {$L_{3}$};
  \node[layer_node,anchor=west,pattern = crosshatch dots] at ([xshift=\base]layer_t_2.east) {};

  \node[layer_node,anchor=west,draw=blue!40,fill=blue!15] (layer_t_4) at ([xshift=2em]layer_t_3.east) {};
  \node[font=\tiny,anchor = north] (layer_t_4_id) at (layer_t_4.south) {$L_{1}$};
  \node[layer_node,anchor=west,pattern = soft horizontal lines] at ([xshift=2em]layer_t_3.east) {};
  \node[layer_node,anchor=west,draw=blue!40,fill=blue!15] (layer_t_5) at ([xshift=\base]layer_t_4.east) {};
  \node[font=\tiny,anchor = north] (layer_t_5_id) at (layer_t_5.south) {$L_{2}$};
  \node[layer_node,anchor=west,pattern =  soft crosshatch] at ([xshift=\base]layer_t_4.east) {};
  \node[layer_node,anchor=west,draw=blue!40,fill=blue!15] (layer_t_6) at ([xshift=\base]layer_t_5.east) {};
  \node[font=\tiny,anchor = north] (layer_t_6_id) at (layer_t_6.south) {$L_{3}$};
  \node[layer_node,anchor=west,pattern = crosshatch dots] at ([xshift=\base]layer_t_5.east) {};
  
  \node[layer_node,anchor=west,draw=red!40,fill=red!20] (layer_t2_1) at ([xshift=11em,yshift = -4em]layer_2_6.east) {};
  \node[font=\tiny,anchor = north] (layer_t2_1_id) at (layer_t2_1.south) {$L_{1}$};
  \node[layer_node,anchor=west,pattern = soft horizontal lines] at ([xshift=11em,yshift = -4em]layer_2_6.east) {};
  \node[layer_node,anchor=west,draw=red!40,fill=red!20] (layer_t2_2) at ([xshift=\base]layer_t2_1.east) {};
  \node[font=\tiny,anchor = north] (layer_t2_2_id) at (layer_t2_2.south) {$L_{2}$};
  \node[layer_node,anchor=west,pattern = soft crosshatch] at ([xshift=\base]layer_t2_1.east) {};
  \node[layer_node,anchor=west,draw=red!40,fill=red!20] (layer_t2_3) at ([xshift=\base]layer_t2_2.east) {};
  \node[font=\tiny,anchor = north] (layer_t2_3_id) at (layer_t2_3.south) {$L_{3}$};
  \node[layer_node,anchor=west,pattern = crosshatch dots] at ([xshift=\base]layer_t2_2.east) {};

  \node[layer_node,anchor=west,draw=blue!40,fill=blue!15] (layer_t2_4) at ([xshift=2em]layer_t2_3.east) {};
  \node[font=\tiny,anchor = north] (layer_t2_4_id) at (layer_t2_4.south) {$L_{1}$};
  \node[layer_node,anchor=west,pattern = soft horizontal lines] at ([xshift=2em]layer_t2_3.east) {};
  \node[layer_node,anchor=west,draw=blue!40,fill=blue!15] (layer_t2_5) at ([xshift=\base]layer_t2_4.east) {};
  \node[font=\tiny,anchor = north] (layer_t2_5_id) at (layer_t2_5.south) {$L_{2}$};
  \node[layer_node,anchor=west,pattern = soft crosshatch] at ([xshift=\base]layer_t2_4.east) {};
  \node[layer_node,anchor=west,draw=blue!40,fill=blue!15] (layer_t2_6) at ([xshift=\base]layer_t2_5.east) {};
  \node[font=\tiny,anchor = north] (layer_t2_6_id) at (layer_t2_6.south) {$L_{3}$};
  \node[layer_node,anchor=west,pattern = crosshatch dots] at ([xshift=\base]layer_t2_5.east) {};

  \node[anchor = west,font=\footnotesize] (more4) at ([xshift = 1.2em,yshift=0.2em]layer_t_6.east) {$\vdots$};
  \node[anchor = west,font=\footnotesize] (more4) at ([xshift = 1.2em,yshift=0.2em]layer_t2_6.east) {$\vdots$};
  \node[draw=ugreen!60,anchor=east,minimum size=1em,fill=ugreen!35,inner sep=0pt] (kd_src) at ([xshift=-1.6em]layer_t_1.west){};
  \node[anchor = east,font=\footnotesize,rotate=-90] (src) at ([xshift = -0.6em,yshift=-2.2em]kd_src.west) {\textsc{Skd}-source};
  \node[draw=red!50,anchor=west,minimum size=1em,fill=red!40,inner sep=0pt] (kd_tgt) at ([xshift=3.2em]layer_t_6.east){};
  \node[anchor = west,font=\footnotesize,rotate=-90] (tgt) at ([xshift = 0.6em,yshift=2.2em]kd_tgt.east) {\textsc{Skd}-target};

  \node[layer_node,anchor=north,draw=red!40,fill=red!20] (student_layer1) at ([yshift=-3em]layer_t2_3.south) {};
  \node[font=\tiny,anchor = north] (layer_3_1_id) at (student_layer1.south) {$L_{1}$};
  \node[layer_node,anchor=north,pattern = soft horizontal lines] at ([yshift=-3em]layer_t2_3.south) {};
  \node[layer_node,anchor=north,draw=blue!40,fill=blue!15] (student_layer2) at ([yshift=-3em]layer_t2_4.south) {};
  \node[font=\tiny,anchor = north] (layer_3_2_id) at (student_layer2.south) {$L_{3}$};
  \node[layer_node,anchor=north,pattern = crosshatch dots] at ([yshift=-3em]layer_t2_4.south) {};
  \node[draw=ugreen!60,anchor=east,minimum size=1em,fill=ugreen!35,inner sep=0pt] at ([xshift=-2.2em,yshift=0.57em]student_layer1.west){};
  \node[draw=red!50,anchor=east,minimum size=1em,fill=red!40,inner sep=0pt] at ([xshift=-2.2em,yshift=-0.57em]student_layer1.west){};
  \node[anchor = east,font=\footnotesize, rotate=-90] (kddata) at ([xshift = -4.2em,yshift=-2em]student_layer1.west) {\textsc{Skd}-Data};
  \node[anchor = west,font=\footnotesize] (more3) at ([xshift = 1.8em,yshift=0.2em]student_layer2.east) {$\vdots$};

  \node[font=\tiny,anchor = north] (space1) at (layer_2.south) {};
  \node[font=\tiny,anchor = north] (space2) at (layer_4.south) {};
  \node[font=\tiny,anchor = north] (space3) at (layer_2_2.south) {};
  \node[font=\tiny,anchor = north] (space4) at (layer_2_4.south) {};
  \node[font=\tiny,anchor = north] (space7) at (layer_n_2.south) {};
  \node[font=\tiny,anchor = north] (space8) at (layer_n_4.south) {};
  \node[font=\tiny,anchor = north] (space9) at (layer_t_2.south) {};
  \node[font=\tiny,anchor = north] (space10) at (layer_t_4.south) {};
    \node[font=\tiny,anchor = north] (space11) at (layer_t2_2.south) {};
  \node[font=\tiny,anchor = north] (space12) at (layer_t2_4.south) {};
 \begin{pgfonlayer}{background}
  \node[draw=black,inner sep=0.4em,rounded corners=4pt,thick,fill=gray!10,drop shadow,minimum width=6.4cm,minimum height=7.4cm] (box1) at (2.0cm, -2.4cm){};
  \node[draw=black,inner sep=0.4em,rounded corners=4pt,thick,fill=gray!10,drop shadow,minimum width=7.2cm,minimum height=2.6cm] (box2) at (9.8cm, 0.0cm){};
  \node[draw=black,inner sep=0.4em,rounded corners=4pt,thick,fill=gray!10,drop shadow,minimum width=7.2cm,minimum height=4cm] (box3) at (9.8cm, -4.1cm){};
  
  \node [draw=black,rectangle,inner sep=0.4em,rounded corners=4pt,drop shadow,fill=orange!10] [fit = (layer_1) (layer_2) (layer_3) (space1)] (group1) {};
  \node [draw=black,rectangle,inner sep=0.4em,rounded corners=4pt,fill=myblue!20,drop shadow] [fit = (layer_4) (layer_5) (layer_6) (space2)] (group2) {};
  \node [draw=black,rectangle,inner sep=0.4em,rounded corners=4pt,fill=orange!10,drop shadow] [fit = (layer_2_1) (layer_2_2) (layer_2_3) (space3)] (group2_1) {};
  \node [draw=black,rectangle,inner sep=0.4em,rounded corners=4pt,line width=0.5pt,fill=myblue!20,drop shadow] [fit = (layer_2_4) (layer_2_5) (layer_2_6) (space4)] (group2_2) {};
  \node [draw=black,rectangle,inner sep=0.4em,rounded corners=4pt,line width=0.5pt,fill=orange!10,drop shadow] [fit = (layer_n_1) (layer_n_2) (layer_n_3) (space7)] (groupn_1) {};
  \node [draw=black,rectangle,inner sep=0.4em,rounded corners=4pt,line width=0.5pt,fill=myblue!20,drop shadow] [fit = (layer_n_4) (layer_n_5) (layer_n_6) (space8)] (groupn_2) {};
  \node [draw=black,rectangle,inner sep=0.4em,rounded corners=4pt,line width=0.5pt,fill=orange!10,drop shadow] [fit = (layer_t_1) (layer_t_2) (layer_t_3) (space9)] (groupt_1) {};
  \node [draw=black,rectangle,inner sep=0.4em,rounded corners=4pt,line width=0.5pt,fill=myblue!20,drop shadow] [fit = (layer_t_4) (layer_t_5) (layer_t_6) (space10)] (groupt_2) {};
  \node [draw=black,rectangle,inner sep=0.4em,rounded corners=4pt,line width=0.5pt,fill=orange!10,drop shadow] [fit = (layer_t2_1) (layer_t2_2) (layer_t2_3) (space11)] (groupt2_1) {};
  \node [draw=black,rectangle,inner sep=0.4em,rounded corners=4pt,line width=0.5pt,fill=myblue!20,drop shadow] [fit = (layer_t2_4) (layer_t2_5) (layer_t2_6) (space12)] (groupt2_2) {};
 \end{pgfonlayer}

	\draw[line width=1pt,draw=red!30,fill=red!20] ([xshift=-0.3em,yshift=-0.4em]group1.south)--([xshift=0.3em,yshift=-0.4em]group1.south) -- ([xshift=0.3em,yshift=-1em]group1.south) -- ([xshift=0.5em,yshift=-1em]group1.south) -- ([xshift=0.0em,yshift=-1.4em]group1.south)-- ([xshift=-0.5em,yshift=-1em]group1.south) -- ([xshift=-0.3em,yshift=-1em]group1.south) -- ([xshift=-0.3em,yshift=-0.4em]group1.south);
	\draw[line width=1pt,draw=blue!35,fill=blue!20] ([xshift=-0.3em,yshift=-0.4em]group2.south)--([xshift=0.3em,yshift=-0.4em]group2.south) -- ([xshift=0.3em,yshift=-1em]group2.south) -- ([xshift=0.5em,yshift=-1em]group2.south) -- ([xshift=0.0em,yshift=-1.4em]group2.south)-- ([xshift=-0.5em,yshift=-1em]group2.south) -- ([xshift=-0.3em,yshift=-1em]group2.south) -- ([xshift=-0.3em,yshift=-0.4em]group2.south);

	\draw[line width=1pt,draw=red!30,fill=red!20] ([xshift=-0.3em,yshift=-0.4em]group2_1.south)--([xshift=0.3em,yshift=-0.4em]group2_1.south) -- ([xshift=0.3em,yshift=-1em]group2_1.south) -- ([xshift=0.5em,yshift=-1em]group2_1.south) -- ([xshift=0.0em,yshift=-1.4em]group2_1.south)-- ([xshift=-0.5em,yshift=-1em]group2_1.south) -- ([xshift=-0.3em,yshift=-1em]group2_1.south) -- ([xshift=-0.3em,yshift=-0.4em]group2_1.south);
	\draw[line width=1pt,draw=blue!35,fill=blue!20] ([xshift=-0.3em,yshift=-0.4em]group2_2.south)--([xshift=0.3em,yshift=-0.4em]group2_2.south) -- ([xshift=0.3em,yshift=-1em]group2_2.south) -- ([xshift=0.5em,yshift=-1em]group2_2.south) -- ([xshift=0.0em,yshift=-1.4em]group2_2.south)-- ([xshift=-0.5em,yshift=-1em]group2_2.south) -- ([xshift=-0.3em,yshift=-1em]group2_2.south) -- ([xshift=-0.3em,yshift=-0.4em]group2_2.south);

	\draw[line width=1pt,draw=red!30,fill=red!20] ([xshift=-0.3em,yshift=-2.8em]group2_1.south)--([xshift=0.3em,yshift=-2.8em]group2_1.south) -- ([xshift=0.3em,yshift=-3.4em]group2_1.south) -- ([xshift=0.5em,yshift=-3.4em]group2_1.south) -- ([xshift=0.0em,yshift=-3.8em]group2_1.south)-- ([xshift=-0.5em,yshift=-3.4em]group2_1.south) -- ([xshift=-0.3em,yshift=-3.4em]group2_1.south) -- ([xshift=-0.3em,yshift=-2.8em]group2_1.south);
	\draw[line width=1pt,draw=blue!35,fill=blue!20] ([xshift=-0.3em,yshift=-2.8em]group2_2.south)--([xshift=0.3em,yshift=-2.8em]group2_2.south) -- ([xshift=0.3em,yshift=-3.4em]group2_2.south) -- ([xshift=0.5em,yshift=-3.4em]group2_2.south) -- ([xshift=0.0em,yshift=-3.8em]group2_2.south)-- ([xshift=-0.5em,yshift=-3.4em]group2_2.south) -- ([xshift=-0.3em,yshift=-3.4em]group2_2.south) -- ([xshift=-0.3em,yshift=-2.8em]group2_2.south);
	\node[anchor=north,inner sep=0pt] at ([yshift=-1.1em]group2_1.south){$\vdots$};
	\node[anchor=north,inner sep=0pt] at ([yshift=-1.1em]group2_2.south){$\vdots$};
  	\node[anchor = south,font=\footnotesize] (group1_text) at (group1.north) {{$\textrm{Group}_{1}$}};
  	\node[anchor = south,font=\footnotesize]  at (group2.north) {$\textrm{Group}_{2}$};
  	\node[anchor = south,font=\footnotesize]  at (groupt_1.north) {{$\textrm{Group}_{1}$}};
  	\node[anchor = south,font=\footnotesize]  at (groupt_2.north) {$\textrm{Group}_{2}$};
  	\node[anchor = south,font=\footnotesize]  at (groupt2_1.north) {{$\textrm{Group}_{1}$}};
  	\node[anchor = south,font=\footnotesize]  at (groupt2_2.north) {$\textrm{Group}_{2}$};
  	\node[anchor = east,font=\footnotesize,rotate=-90] (batch1) at ([xshift = -2em,yshift=-1.6em]layer_1.west) {$\textrm{Batch}_{1}$};
  	\node[anchor = east,font=\footnotesize,rotate=-90] (batch2) at ([xshift = -2em,yshift=-1.6em]layer_2_1.west) {$\textrm{Batch}_{2}$};
  	\node[anchor = east,font=\footnotesize,rotate=-90] (batchn) at ([xshift = -2em,yshift=-1.6em]layer_n_1.west) {$\textrm{Batch}_{n}$};

  	\node[anchor = north,font=\scriptsize] (reodrer) at([xshift=3em,yshift=-0.2em]group1.south) {reorder};

  	\node[anchor = west] (more1) (more1) at ([xshift = 1.5em,yshift=0.2em]layer_6.east) {$\vdots$};
  	\node[anchor = west] (more2) at ([xshift = 1.5em,yshift=0.2em]layer_2_6.east) {$\vdots$};
  	\node[anchor = west] (moren) at ([xshift = 1.5em,yshift=0.2em]layer_n_6.east) {$\vdots$};

  \draw[-latex',thick] ([xshift=-1.5em]layer_1.west) -- (layer_1.west);
  \draw[-latex',thick] (layer_1.east) -- (layer_2.west);
  \draw[-latex',thick] (layer_2.east) -- (layer_3.west);
  \draw[-latex',thick] (layer_3.east) -- (layer_4.west);
  \draw[-latex',thick] (layer_4.east) -- (layer_5.west);
  \draw[-latex',thick] (layer_5.east) -- (layer_6.west);
  \draw[-latex',thick] (layer_6.east) -- ([xshift=1.5em]layer_6.east);

  \draw[-latex',thick] ([xshift=-1.5em]layer_2_1.west) -- (layer_2_1.west);
  \draw[-latex',thick] (layer_2_1.east) -- (layer_2_2.west);
  \draw[-latex',thick] (layer_2_2.east) -- (layer_2_3.west);
  \draw[-latex',thick] (layer_2_3.east) -- (layer_2_4.west);
  \draw[-latex',thick] (layer_2_4.east) -- (layer_2_5.west);
  \draw[-latex',thick] (layer_2_5.east) -- (layer_2_6.west);
  \draw[-latex',thick] (layer_2_6.east) -- ([xshift=1.5em]layer_2_6.east);

  \draw[-latex',thick] ([xshift=-1.5em]layer_n_1.west) -- (layer_n_1.west);
  \draw[-latex',thick] (layer_n_1.east) -- (layer_n_2.west);
  \draw[-latex',thick] (layer_n_2.east) -- (layer_n_3.west);
  \draw[-latex',thick] (layer_n_3.east) -- (layer_n_4.west);
  \draw[-latex',thick] (layer_n_4.east) -- (layer_n_5.west);
  \draw[-latex',thick] (layer_n_5.east) -- (layer_n_6.west);
  \draw[-latex',thick] (layer_n_6.east) -- ([xshift=1.5em]layer_n_6.east);

  \draw[-latex',thick] ([xshift=-1.5em]student_layer1.west) -- (student_layer1.west);
  \draw[-latex',thick] (student_layer1.east) -- (student_layer2.west);
  \draw[-latex',thick] (student_layer2.east) -- ([xshift=1.5em]student_layer2.east);
  
  \draw[-latex',thick] ([xshift=-1.5em]layer_t_1.west) -- (layer_t_1.west);
  \draw[-latex',thick] (layer_t_1.east) -- (layer_t_2.west);
  \draw[-latex',thick] (layer_t_2.east) -- (layer_t_3.west);
  \draw[-latex',thick] (layer_t_3.east) -- (layer_t_4.west);
  \draw[-latex',thick] (layer_t_4.east) -- (layer_t_5.west);
  \draw[-latex',thick] (layer_t_5.east) -- (layer_t_6.west);
  \draw[-latex',thick] (layer_t_6.east) -- ([xshift=1.5em]layer_t_6.east);
  \draw[-latex',thick] ([xshift=1.9em]layer_t_6.east) -- ([xshift=3.0em]layer_t_6.east);

  \draw[-latex',very thick,red!40] (groupt2_1.south)..controls +(south:1em) and +(north:1.5em)..(student_layer1.north);
  \draw[-latex',very thick,blue!40] (groupt2_2.south)..controls +(south:1em) and +(north:1.5em)..(student_layer2.north);
	\node[font=\tiny] (sampling) at ([xshift = -1.5em,yshift = 0.8em]student_layer2.north) {sampling};

  \node [auto,anchor=west,font=\footnotesize,rotate=-90] at ([xshift=-1em,yshift=3em]box1.east){Teacher Training} ;

  \node [auto,anchor=west,font=\footnotesize,rotate=-90] at ([xshift=-2em,yshift=3em]box3.east){Student Training} ;
  
   \node[auto,anchor=south,font=\footnotesize,inner sep=0pt] at ([xshift =-6.4em,yshift = -1em]box2.north) {Generate \textsc{Skd}-data} ;
  
  \node[draw=gray!70,line width=1pt,fill=gray!10,single arrow,minimum height=2.2em,minimum width=4pt,single arrow head extend=3pt] (kd) at ([xshift=-1.4em]box2.west){};
   \node[draw=gray!70,line width=1pt,fill=gray!10,single arrow,minimum height=1.6em,minimum width=4pt,single arrow head extend=3pt,rotate=-90] (kd) at ([yshift=1.15em]box3.north){};

  \end{tikzpicture}
   \caption{An overview of the \textsc{Gpkd} method including three stages. $\mathrm{Group}_{1}$ and $\mathrm{Group}_{2}$ correspond to different groups of the stacking layers. $L_{1}$, $L_{2}$ and $L_{3}$ denote the layers in each group.} \label{fig:compression}
 \end{figure*}
 
We ran experiments on the WMT16 English-German, NIST OpenMT12 Chinese-English and WMT19 Chinese-English translation tasks. The \textsc{Gpkd} method compressed a 48-layer Transformer into a 6-layer system with almost no loss in BLEU. It outperformed the baseline with the same depth by +$2.46$ BLEU points. Through skipping sub-layer method, the teacher network achieved a BLEU score of $30.63$ BLEU on the \textit{newstest2014} English-German task, and the student obtains additional improvements of $0.50$ BLEU points.


\section{Compression of Deep Transformer}

In this section, we first introduce the formulation of knowledge distillation (\textsc{Kd}), then present the group-permutation based knowledge distillation (\textsc{Gpkd}) approach to compressing deep Transformer.

\subsection{Knowledge Distillation}

The purpose of \textsc{Kd} is to transfer knowledge from a complex teacher network to a light-weight student network by encouraging the student network reproducing the performance of the teacher network. Let $\mathcal{F}^{T}(x,y)$ and $\mathcal{F}^{S}(x,y)$ represent the predictions of the teacher network and the student network, respectively. Then \textsc{Kd} can be formulated as follows:
\begin{equation}
  \mathcal{L}_{\textsc{Kd}} = \sum_{x \in \mathcal{X}, y \in \mathcal{Y}} L\left(\mathcal{F}^{T}(x,y), \mathcal{F}^{S}(x,y)\right)
\end{equation}
\noindent where $L(\cdot)$ is a loss function to evaluate the distance between $\mathcal{F}^{T}(x,y)$ and $\mathcal{F}^{S}(x,y)$. $x$ and $y$ represent the source inputs and target inputs, respectively. $(\mathcal{X}$,$\mathcal{Y})$ denotes the whole training dataset. The objective can  be seen as minimizing the loss function to bridge the gap between the student and its teacher.

To advance the student model, a promising method is to learn from the intermediate layer representations of the teacher network via additional loss functions \cite{sun-etal-2019-patient, jiao2019tinybert}. However, the additional loss functions require large memory footprint due to the logits computation of both the teacher and student networks in each mini-batch training phase. This is quite challenging when the teacher is extremely deep. Alternatively, we choose sequence-level knowledge distillation (\textsc{Skd}), proposed in \citet{kim-rush-2016-sequence}'s work to simplify the training procedure. Through their results, \textsc{Skd} achieves comparable or even higher translation performance than word-level \textsc{Kd} method. Concretely, \textsc{Skd} uses the translation results of the teacher network as the gold instead of the ground truth. In this work, we build our student systems upon \textsc{Skd} method.


\subsection{Group-Permutation Based Knowledge Distillation}

Through preliminary experimental results, the student network trained with the \textsc{Skd} data still significantly underperforms its teacher. A possible explanation is directly shrinking the encoder depth is harmful to the translation performance. To further bridge the gap between the teacher network and the student work, we propose a group-permutation based knowledge distillation method, including three stages: (\romannumeral1) group-permutation training strategy which rectifies the information flow of the teacher network during training phase. (\romannumeral2) generate the \textsc{Skd} data through the teacher network. (\romannumeral3) train the student network with the \textsc{Skd} data. Instead of randomly initializing the parameters of the student network, we selected layers from the teacher network to form the student network, which provides a better initialization. Figure 1 exhibits the whole training process.

\subsubsection{Group-Permutation Training}
Assuming that the teacher model has $N$ Transformer layers, we aim to extract $M$ layers to form a student model. This can be characterized as learning the mapping between $layer_{m}$ and $layer_{n}$, where $m \in \{1,2...,M\}$ and $n=N/M * m$. To achieve this goal, the stacking layers are first divided into $M$ groups and every adjacent  $h=N/M$ layers form a group. The core idea of the proposed method is to make the selected single layer mimic the behavior of its group output. Instead of employing additional loss functions to reduce the distance of the intermediate layer representations between the student network and the teacher network, we simply disturbing the computation order in each group when training the teacher network.

In each mini-batch of training, we randomly disturb the order of the layers in each group. Suppose that $G_{i}=\{L_{1},L_{2},L_{3}\} (i \in M)$  is the set of intermediate layers in each group. The left part of Figure \ref{fig:compression} shows the computation order of the layers during the teacher training phase. Note that each group is independent with others in ordering the layers. In this work, we sample the layer order uniformly from all the permutation choices\footnote{For a 3-layer group, the computation order includes $\{L_{1},L_{2},L_{3}\}$, $\{L_{1},L_{3},L_{2}\}$, $\{L_{2},L_{1},L_{3}\}$, $\{L_{2},L_{3},L_{1}\}$, $\{L_{3},L_{1},L_{2}\}$ and $\{L_{3},L_{2},L_{1}\}$}.

\subsubsection{Generating \textsc{Skd} Data}
As shown in Stage 2 (the top right part in Figure \ref{fig:compression}), given the training dataset $\{\mathcal{X},\mathcal{Y}\}$, the teacher network translates the source inputs into the target sentences $\mathcal{Z}$. Then the \textsc{Skd} data is the collection of $\{\mathcal{X},\mathcal{Z}\}$.

\subsubsection{Student Training}

After obtaining the teacher network and the \textsc{Skd} data, we begin optimizing the student network with the supervision of the teacher. As illustrated in Stage 1, each single layer imitates the logits of its group output and  every layer in each group behave similar with each other. Then we randomly choose one layer from each group to form a ``new'' encoder which the encoder depth is reduced from $N$ to $M$. It can be regarded as the student model with fewer layers. The compression rate is controlled by the hyper-parameter $h$. One can compress a 48-layer teacher into a 6-layer network by setting $h=8$, or progressively achieve this goal with $h=2$ for three times of compression.

\begin{figure}[t]
  \small
  \centering
  \centering
  \begin{tikzpicture}{
    \node [] (Base) at (0,0) {};
    \node [] (Base48) at ([xshift=1em]Base.east) {\small{Training}};
    \node [] (SDT-48L) at ([xshift=5em]Base48.east) {\small{Validation}};
    \draw[red,mark=otimes*,line width=0.75pt] ([xshift=0.1em]Base48.east) -- plot[mark=diamond*]([xshift=1.1em]Base48.east) -- ([xshift=2.1em]Base48.east);
    \draw[blue,mark=triangle*,line width=0.75pt] ([xshift=0.1em]SDT-48L.east) -- plot[mark=triangle*]([xshift=1.1em]SDT-48L.east) -- ([xshift=2.1em]SDT-48L.east);
  }
  \end{tikzpicture}
   \\
   \centering
    \begin{tikzpicture}
      \begin{axis}[
      width=.22\textwidth,
      height=.20\textwidth,
      legend style={at={(0.63,0.61)}, anchor=south west},
      xlabel={\scriptsize{Epoch}},
      ylabel={\scriptsize{Training PPL}},
      ylabel style={yshift=-1.7em},xlabel style={yshift=0.7em},
      yticklabel style={/pgf/number format/precision=1,/pgf/number format/fixed zerofill},
      axis y line*=left,
      ymin=5.7,ymax=7.3, ytick={6.0, 6.5, 7.0},
      xmin=14.5,xmax=25.5,xtick={15, 17, 19, 21, 23, 25},
      legend style={at={(0.333,0.01)}, anchor=south west, legend plot pos=right,font=\tiny,cells={anchor=west}}
      ]

      \addplot[red,mark=diamond*,line width=0.75pt] coordinates {(13,6.86) (14,6.80) (15,6.76) (16,6.72) (17,6.68) (18,6.64) (19,6.61) (20,6.59) (21,6.59) (22,6.59) (23,6.58) (24,6.59) (25, 6.59)};
      \end{axis}
      \begin{axis}[
        width=.22\textwidth,
        height=.20\textwidth,
        xlabel={\scriptsize{Epoch}},
        ylabel={\scriptsize{Validation PPL}},
        ylabel style={yshift=-13.4em},xlabel style={yshift=0.7em},
        yticklabel style={/pgf/number format/precision=1,/pgf/number format/fixed zerofill},
        axis y line*=right,
        ymin=3.67,ymax=3.93, ytick={3.7,3.8,3.9},
        xmin=14.5,xmax=25.5,xtick={15, 17, 19, 21, 23, 25},
       ]
        \addplot[blue,mark=triangle*,line width=0.75pt] coordinates {(13,3.85) (14,3.85) (15,3.83) (16,3.78) (17,3.75) (18,3.76) (19,3.75) (20,3.73) (21,3.74) (22,3.73) (23,3.74) (24,3.74) (25, 3.74)};
      \end{axis}
      \end{tikzpicture}
    \centering
    \begin{tikzpicture}
        \begin{axis}[
          width=.22\textwidth,
        height=.20\textwidth,
        legend style={at={(0.63,0.61)}, anchor=south west},
        xlabel={\scriptsize{Epoch}},
        ylabel={\scriptsize{Training PPL}},
        ylabel style={yshift=-1.7em},xlabel style={yshift=0.7em},
        yticklabel style={/pgf/number format/precision=1,/pgf/number format/fixed zerofill},
        axis y line*=left,
        ymin=4.5,ymax=6.5, ytick={5.0, 5.5, 6.0},
        xmin=14.5,xmax=25.5,xtick={15, 17, 19, 21, 23, 25},
        legend style={at={(0.333,0.01)}, anchor=south west, legend plot pos=right,font=\tiny,cells={anchor=west}}
        ]

        \addplot[red,mark=diamond*,line width=0.75pt] coordinates {(13,6.00) (14,5.89) (15,5.80) (16,5.71) (17,5.64) (18,5.57) (19,5.51) (20,5.45) (21,5.40) (22,5.38) (23,5.35) (24,5.31) (25, 5.29)};
        \end{axis}
        \begin{axis}[
          width=.22\textwidth,
          height=.20\textwidth,
          legend style={at={(0.63,0.61)}, anchor=south west},
          xlabel={\scriptsize{Epoch}},
          ylabel={\scriptsize{Validation PPL}},
          ylabel style={yshift=-13.4em},xlabel style={yshift=0.7em},
          yticklabel style={/pgf/number format/precision=1,/pgf/number format/fixed zerofill},
          axis y line*=right,
          ymin=3.47,ymax=3.73, ytick={3.5,3.6,3.7},
          xmin=14.5,xmax=25.5,xtick={15, 17, 19, 21, 23, 25},
          legend style={at={(0.333,0.01)}, anchor=south west,
          legend plot pos=right,font=\tiny,cells={anchor=west}}
         ]
          \addplot[blue,mark=triangle*,line width=0.75pt] coordinates {(13,3.54) (14,3.54) (15,3.53) (16,3.52) (17,3.52) (18,3.54) (19,3.52) (20,3.53) (21,3.53) (22,3.55) (23,3.53) (24,3.56) (25, 3.60)};
        \end{axis}
    \end{tikzpicture}
  \caption{The comparison of training and validation PPL on the shallow (left) and the deep (right) models.}
  \label{fig:train-valid ppl}

\end{figure}
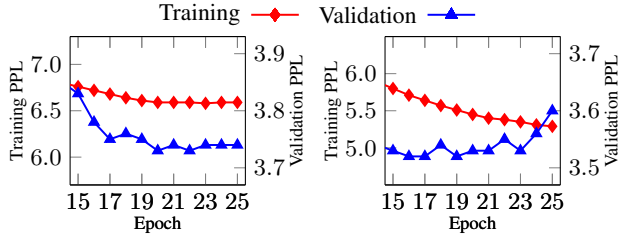

\subsection{The \textsc{Desdar} Architecture}

The \textsc{Gpkd} method also fits into the decoder. Hence we design a heterogeneous NMT model consisting of a deep encoder and a shallow decoder, abbreviated as (\textsc{Desdar}). Such an architecture can enjoy high translation quality due to the deep encoder and fast inference due to the light decoder. Similar findings were observed in previous work \cite{zhang-etal-2019-improving,xiao2019sharing}, which could speed up the Transformer inference by simplifying the decoder architecture. This is due to the fact that the heavy use of dot-product attention in the decoder and the nature of auto-regressive decoding slows down the system. 

Moreover, it offers a way of balancing the translation quality and the inference. This is promising for industrial applications. For example, one can maximize the speedup by using one decoder layer or two, or can yield further BLEU improvements by enlarging the decoder depth. The experimental results in the following sections show the effectiveness of the \textsc{Desdar} architecture.

\section{Skipping Sub-Layers for Deep Transformer}

It is well known that stronger teacher networks (deeper or wider) can bring better supervision signals \cite{hinton2015distilling}. However, these over-parameterized networks often face the overfitting problem \cite{fan2019reducing}. In this section, we introduce a regularization training method to alleviate the overfitting problem which can further improve the extremely deep Transformer systems. Here we start with the sub-layer co-adaptation problem, followed by the \textit{Skipping Sub-Layer} method to address it. 


\subsection{Co-adaptation of Sub-Layers}

In Transformer, a stack of sub-layers (or layers) are used between the input and the output. The relationship between the input and the output is complex if the model goes deeper. There will typically be co-adaptation of the sub-layers, and a sub-layer will operate based on the states of other sub-layers. This property is helpful for training because sub-layers are learned to work well together on the training data. But it prevents the model from generalizing well at test time. This is similar to the case in that too many hidden units of a layer lead to the co-adaptation of feature detectors \cite{hinton2012improving}. See Figure \ref{fig:train-valid ppl} for perplexities on the training and validation data at different training steps. Clearly, the deep model (48-layer encoder) appears to overfit.

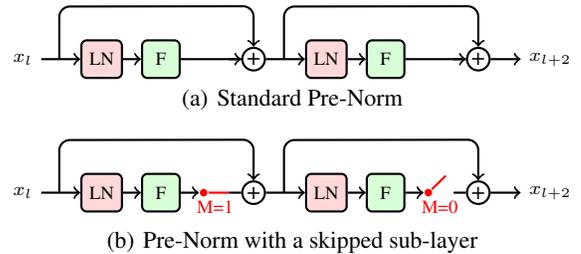
\begin{figure}
  \small
  \centering
  \subfigure[Standard Pre-Norm]{
  \centering
    \begin{tikzpicture}
      \tikzstyle{sublayernode} = [rectangle,draw,thick,inner sep=3pt,rounded corners=2pt,align=center,minimum height=0.5cm,minimum width=0.5cm,font=\scriptsize]
      \tikzstyle{inputnode} = [rectangle,inner sep=3pt,align=center,font=\scriptsize]
      \tikzstyle{circlenode} = [circle,draw,thick,minimum size=0.3\base,font=\small,inner sep=0pt]
      \tikzstyle{mnode} = [circle,thick,minimum size=0.1\base,font=\small,inner sep=0pt]

      \node[anchor=south west,inputnode] (input) at (0,0) {$x_{l}$};

      \node[anchor=west,sublayernode,fill=red!15] (ln) at ([xshift=1.3\hseg]input.east) {LN};
      \node[anchor=west,sublayernode,fill=green!15] (fn) at ([xshift=0.6\hseg]ln.east) {F};
      \node[anchor=west,mnode] (m) at ([xshift=0.6\hseg]fn.east) {};
      \node[circlenode] (res) at ([xshift=1.5\hseg]m.east) {+};

      \node[anchor=west,sublayernode,fill=red!15] (ln1) at ([xshift=1.3\hseg]res.east) {LN};
      \node[anchor=west,sublayernode,fill=green!15] (fn1) at ([xshift=0.6\hseg]ln1.east) {F};
      \node[anchor=west,mnode] (m1) at ([xshift=0.6\hseg]fn1.east) {};
      \node[circlenode] (res1) at ([xshift=1.5\hseg]m1.east) {+};

      \node[anchor=west,inputnode] (output) at ([xshift=1.0\hseg]res1.east) {$x_{l+2}$};



      \coordinate (mend) at ([xshift=1.0\hseg]m.west);
      \draw[->,thick] (input)--(ln);
      \draw[->,thick] (ln)--(fn);
      \draw[->,thick] (fn)--(res);
      \draw[-,thick] (mend)--(res);
      \coordinate (h) at ([xshift=-0.7\hseg]ln.west);
      \draw[->,thick,rounded corners] (h) -- ([shift={(0,0.40\wnode-0.2\wseg)}]h.north) -- ([shift={(0,0.36\wseg)}]res.north) -- (res.north);

      \coordinate (mend1) at ([xshift=1.0\hseg]m1.west);
      \draw[->,thick] (res)--(ln1);
      \draw[->,thick] (ln1)--(fn1);
      \draw[->,thick] (fn1)--(res1);
      \draw[-,thick] (mend1)--(res1);
      \draw[->,thick] (res1)--(output);
      \coordinate (h1) at ([xshift=-0.7\hseg]ln1.west);
      \draw[->,thick,rounded corners] (h1) -- ([shift={(0,0.40\wnode-0.2\wseg)}]h1.north) -- ([shift={(0,0.36\wseg)}]res1.north) -- (res1.north);
    \end{tikzpicture}
  }
 
  \subfigure[Pre-Norm with a skipped sub-layer]{
   \centering
    \begin{tikzpicture}
      \tikzstyle{sublayernode} = [rectangle,draw,thick,inner sep=3pt,rounded corners=2pt,align=center,minimum height=0.5cm,minimum width=0.5cm,font=\scriptsize]
      \tikzstyle{inputnode} = [rectangle,inner sep=3pt,align=center,font=\scriptsize]
      \tikzstyle{circlenode} = [circle,draw,thick,minimum size=0.3\base,font=\small,inner sep=0pt]
      \tikzstyle{mnode} = [circle,thick,minimum size=0.1\base,fill=red,font=\small,inner sep=0pt]

      \node[anchor=south west,inputnode] (input) at (0,0) {$x_{l}$};

      \node[anchor=west,sublayernode,fill=red!15] (ln) at ([xshift=1.3\hseg]input.east) {LN};
      \node[anchor=west,sublayernode,fill=green!15] (fn) at ([xshift=0.6\hseg]ln.east) {F};
      \node[anchor=west,mnode] (m) at ([xshift=0.6\hseg]fn.east) {};
      \node[circlenode] (res) at ([xshift=1.5\hseg]m.east) {+};

      \node[anchor=west,sublayernode,fill=red!15] (ln1) at ([xshift=1.3\hseg]res.east) {LN};
      \node[anchor=west,sublayernode,fill=green!15] (fn1) at ([xshift=0.6\hseg]ln1.east) {F};
      \node[anchor=west,mnode] (m1) at ([xshift=0.6\hseg]fn1.east) {};
      \node[circlenode] (res1) at ([xshift=1.5\hseg]m1.east) {+};

      \node[anchor=west,inputnode] (output) at ([xshift=1.0\hseg]res1.east) {$x_{l+2}$};

      \node[anchor=south west,inputnode,red] (mlable) at ([shift={(-0.5\hseg,-0.8\hseg)}]m.south) {M=1};

      \node[anchor=south west,inputnode,red] (mlable1) at ([shift={(-0.5\hseg,-0.8\hseg)}]m1.south) {M=0};

      \coordinate (mend) at ([xshift=1.0\hseg]m.west);
      \draw[->,thick] (input)--(ln);
      \draw[->,thick] (ln)--(fn);
      \draw[->,thick] (fn)--(m);
      \draw[-,thick,red] (m)--(mend);
      \draw[-,thick] (mend)--(res);
      \coordinate (h) at ([xshift=-0.7\hseg]ln.west);
      \draw[->,thick,rounded corners] (h) -- ([shift={(0,0.40\wnode-0.2\wseg)}]h.north) -- ([shift={(0,0.36\wseg)}]res.north) -- (res.north);

      \coordinate (mend1) at ([xshift=1.0\hseg]m1.west);
      \draw[->,thick] (res)--(ln1);
      \draw[->,thick] (ln1)--(fn1);
      \draw[->,thick] (fn1)--(m1);
      \draw[-,thick,red] (m1)--([xshift=-0.07\wseg,yshift=0.15\wseg]mend1);
      \draw[-,thick] (mend1)--(res1);
      \draw[->,thick] (res1)--(output);
      \coordinate (h1) at ([xshift=-0.7\hseg]ln1.west);
      \draw[->,thick,rounded corners] (h1) -- ([shift={(0,0.40\wnode-0.2\wseg)}]h1.north) -- ([shift={(0,0.36\wseg)}]res1.north) -- (res1.north);

    \end{tikzpicture}
  }
  \caption{The sub-layer information flow of (a) Standard Pre-Norm, (b) Pre-Norm with a skipped bub-layer.}
  \label{fig:arch}
  \end{figure}

\subsection{The Skipping Sub-Layer Method}

To address the overfitting problem, we drop either the self-attention sub-layer or the feed-forward sub-layer of the Transformer encoder for robust training (call it the \textit{Skipping Sub-Layer} method). Both types of the sub-layer follow the Pre-Norm architecture of deep Transformer \cite{wang-etal-2019-learning}:

\begin{equation}
  x_{l+1} =  \textrm{F}(\mathrm{LN}(x_{l})) + x_{l} \label{eq:standard}
\end{equation}

\noindent where $\mathrm{LN(\cdot)}$ is the layer normalization function, $x_l$ is the output of sub-layer $l$ and $\mathrm{F(\cdot)}$ is either the self-attention or feed-forward function. We use variable $M$ $\in$ $\{0,1\}$ to  control how often a sub-layer is omitted. Then, we re-define the sub-layer as:

\begin{equation}
  x_{l+1} =  M \cdot \textrm{F}(\mathrm{LN}(x_{l})) + x_{l}
\end{equation}

\noindent where $M=0\ (\textrm{or}=1)$ means that the sub-layer is omitted (or reserved). See Figure \ref{fig:arch} for comparison of two sub-layer networks. 

In addition, \citet{iclrGreff2017} have shown that the lower-level sub-layers of a deep neural network provide the core representation of the input and the subsequent sub-layers refine that representation. Therefore, it is natural to skip fewer sub-layers if they are close to the input, instead of using the same dropping rate for each single layer in \citet{fan2019reducing}. To this end, we design a method that makes the lower-level sub-layers seldom to be dropped. Let $L$ be the number of layers of the stack\footnote{There are $2L$ sub-layers for $L$ layers.} and $l$ be the current sub-layer. Then, we define $M$ as

\begin{equation}
  M=\left\{
    \begin{array}{ll}
    0, & P \leq p_{l} \\
    1, & P > p_{l}
  \end{array}
  \right.
  \end{equation}

\noindent where

\begin{equation}
  p_{l}=\frac{l}{2L} \cdot \phi,\ \ \ \ \ \textrm{for} \ 1\leq l \leq 2L
  \end{equation}

\noindent In this model, $\phi$ is a hyper-parameter that is set to 0.4 in our experiments. $p_l$ is the rate of omitting the sub-layer. For sub-layer $l$, we first draw a variable $P$ from the uniform distribution in $[0,1]$. Then, $M$ is set to 1 if $P > p_{l}$, and 0 otherwise. Thus the lower-level sub-layers are more likely to survive.

Note that the \textit{Skipping Sub-Layer} method is doing something like sampling a sub-network from a full network. For a model with $2L$ sub-layers, it encodes $2^{2L}$ sub-networks and each configuration of sub-layer omission represents a sub-network. These sub-models are learned efficiently because they share the parameters. For inference, all these sub-models behave like an ensemble model. Following the work in \citet{hinton2012improving}, we rescale the output representation of each sub-layer by the survival rate $1-p_{l}$, like this:

\begin{equation}
  x_{l+1} =  (1-p_{l}) \cdot \textrm{F}(\mathrm{LN}(x_{l})) + x_{l}
\end{equation}
\noindent Factor $1-p_{l}$ is used to scale-down the output of the sub-layer, so the expected output of the sub-layer is the same as the actual output at test time. Then, the final model can make a more accurate prediction by averaging the predictions from $2^{2L}$ sub-models.

\subsection{Two-stage Training}
\label{sec:two-state-training}

Our \textit{Skipping Sub-Layer} method is straightforwardly applicable to the training phase of Transformer. However, we found in our preliminary experiments that the learned model even underperformed the baseline if we introduced sub-layer omission into training from the beginning. This might be due to the fact that deep Transformer is complex and the training is fragile to the perturbation if the model does not get to the smoothed region of the error surface.

Here, we instead adopt a two-stage training method to learn the deep Transformer model with omitting sub-layers. First, we train the model as usual but early stop it when the model converges on the validation set. Then, we apply our \textit{Skipping Sub-Layer} method to the model and continue training until the model converges again. As is shown in Table 4, the two-stage training is helpful for making better use of random sub-layer omission and producing better results. To our knowledge, we are the first to emphasize the importance of the two-stage training in building deep Transformer with omitting layers or sub-layers.

\section{Experiments}

We conducted experiments on the WMT'16 English-German, NIST'12 Chinese-English and WMT19' Chinese-English tasks.

\begin{table*}[t]
  \small
  \setlength{\tabcolsep}{1.5pt}
  \centering
  \begin{tabular}{l r r c c r c c c c c r c c c c c}
    \toprule
    \multirow{2}{*}{\textbf{Systems}} & \multirow{2}{*}{\textbf{Depth}} & \multicolumn{3}{c }{\textbf{WMT En-De}} & \multicolumn{6}{c}{\textbf{NIST Zh-En}} &\multicolumn{6}{c}{\textbf{WMT Zh-En}} \\
    \cmidrule(lr){3-5} \cmidrule(lr){6-11} \cmidrule(lr){12-17}
    & & \textbf{Params} & \textbf{BLEU} &\textbf{$\Delta$} & \textbf{Params} & \textbf{MT06} & \textbf{MT04} & \textbf{MT08} &\textbf{Avg} &\textbf{$\Delta$} & \textbf{Params} & \textbf{Test17} & \textbf{Test18} & \textbf{Test19} &\textbf{Avg} &\textbf{$\Delta$}\\
    \midrule
    Deep-RPR-24L  &24-6  &118M  &29.39  &-       &141M  &53.56  &56.31  &48.15   &52.67  &-      &142M  &26.50  &27.00  &28.30   &27.26  &-        \\
    Base-RPR-6L   &6-6   &62M   &27.60  &ref     &84M   &51.63  &55.06  &46.17   &50.95  &ref    &85M   &25.90  &26.00  &27.60   &26.50  &ref      \\
    +\textsc{Skd} &6-6   &62M   &28.50  &+0.90   &84M   &52.60  &55.16  &46.99   &51.58  &+0.63  &85M   &26.10  &26.20  &27.80   &26.70  &+0.20    \\
    +\textsc{Gpkd}&6-6   &62M   &29.33  &+1.73   &84M   &53.32  &56.24  &47.65   &52.40  &+1.45  &85M   &26.40  &26.90  &28.30   &27.20  &+0.70    \\
    \midrule            
    Deep-RPR-48L  &48-6  &194M  &30.03  &-       &217M  &54.00  &56.40  &48.21   &52.87  &-      &218M  &26.80  &27.30  &28.60   &27.56  &-        \\
    Base-RPR-6L   &6-6   &62M   &27.60  &ref     &84M   &51.63  &55.06  &46.17   &50.95  &ref    &85M   &25.90  &26.00  &27.60   &26.50  &ref      \\
    +\textsc{Skd} &6-6   &62M   &29.01  &+1.39   &84M   &52.55  &55.15  &46.92   &51.54  &+0.59  &85M   &26.40  &26.40  &28.00   &26.93  &+0.43    \\ 
    +\textsc{Gpkd}&6-6   &62M   &29.68  &+2.08   &84M   &53.63  &56.19  &48.11   &52.64  &+1.69  &85M   &26.70  &27.10  &28.50   &27.43  &+0.93    \\
    \midrule            
    Big-RPR-12L   &12-6  &286M  &29.91  &-       &332M  &54.19  &56.89  &49.29   &53.45  &-      &335M  &27.00  &27.50  &28.70   &27.73  &-        \\
    Big-RPR-6L    &6-6   &211M  &29.21  &ref     &256M  &52.80  &55.57  &47.54   &51.97  &ref    &258M  &26.50  &27.00  &28.20   &27.23  &ref      \\
    +\textsc{Skd} &6-6   &211M  &29.47  &+0.26   &256M  &53.73  &55.80  &47.78   &52.43  &+0.46  &258M  &26.80  &27.30  &28.50   &27.53  &+0.30    \\ 
    +\textsc{Gpkd}&6-6   &211M  &29.88  &+0.67   &256M  &54.03  &56.55  &49.16   &53.24  &+1.27  &258M   &26.90  &27.40  &28.70   &27.66  &+0.43    \\
    \bottomrule     
  \end{tabular}
  \caption{The results of the \textsc{Gpkd} method applied in encoder side on En-De and Zh-En tasks. We set $h=4$ and $h=8$ to compress the 24-layer and 48-layer systems, respectively. RPR denotes the Transformer incorporating byrelative positional information.}
  \label{tab:compression}
\end{table*}

\subsection{Experimental Setups}
The bilingual and evaluation data mainly came from three sources:

\begin{itemize}
  \item WMT'16 English-German (En-De). We used the same datasets as in \cite{vaswani2017attention,wu2019pay,wang-etal-2019-learning}. They consisted of approximately $4.5$M tokenized sentence pairs. All sentences were segmented into sequences of sub-word units \cite{sennrich-subword-neural} with $32$K merge operations using a vocabulary shared by the source and target sides. \textit{newstest2016} and \textit{newstest2014} was the validation and test data, respectively.
  \item NIST'12 Chinese-English (NIST Zh-En). We randomly extracted nearly 1.9M bilingual corpus from NIST'12 OpenMT\footnote{LDC Number: LDC2000T46, LDC2000T47, LDC2000T50, LDC2003E14, LDC2005T10, LDC2002E18, LDC2007T09, and LDC2004T08}. MT06 was the validation set and the concatenation of MT04 and MT08 was the test set. 
  \item WMT'19 Chinese-English (WMT Zh-En). For more convincing results, we also experimented on a larger dataset extracted from the mixture of the CWMT and UN corpora, provided by \citet{wang-etal-2019-learning}. We selected \textit{newstest2017} as the validation data and reported the BLEU scores on \textit{newstest2018} and \textit{newstest2019}.
\end{itemize}

We adopted the compound split strategy for En-De, which was a common post-processing step used
in previous work \cite{vaswani2017attention,wang-etal-2019-learning,wu-etal-2019-depth}. For Zh-En tasks, all the sentences were segmented by the tool provided within NiuTrans \cite{xiao2012niutrans}. Translation quality was measured by case-sensitive tokenized BLEU for En-De task, and case-insensitive tokenized BLEU for NIST Zh-En task. The BLEU script was \textsl{multi-bleu.perl}. We also reported the sacrebleu\footnote{BLEU+case.mixed+numrefs.1+smooth.exp+tok.13a\\+version.1.2.12} results on the En-De and the WMT Zh-En tasks, respectively. 
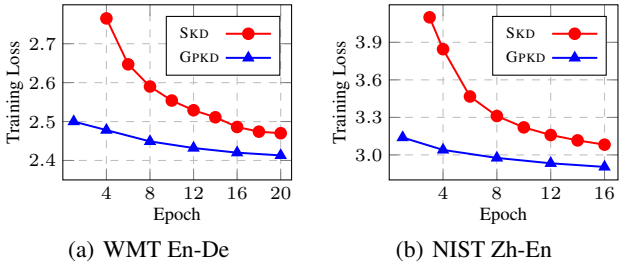
\begin{figure}[t]

  \small
  \subfigure[WMT En-De]
  {
      \centering
      \begin{tikzpicture}
      \scriptsize{
          \begin{axis}[
            ymajorgrids,
          xmajorgrids,
          grid style=dashed,
          width=.26\textwidth,
          height=.22\textwidth,
          xlabel={\scriptsize{Epoch}},
          ylabel={\scriptsize{Training Loss}},
          ylabel style={yshift=-2.4em},xlabel style={yshift=1.0em},
          ymin=2.35,ymax=2.8, ytick={ 2.4,2.5,2.6, 2.7},
          yticklabel style={/pgf/number format/precision=1,/pgf/number format/fixed zerofill},
          xmin=0,xmax=21,xtick={4, 8, 12, 16, 20},
          legend style={at={(0.45,0.595)}, anchor=south west, legend plot pos=right,font=\tiny,cells={anchor=west}}
          ]

          \addplot[red,mark=otimes*,line width=0.75pt] coordinates {(4,2.765) (6,2.647) (8,2.590) (10,2.554) (12,2.529) (14,2.511) (16,2.486) (18,2.474) (20,2.470)};
          \addlegendentry{\textsc{Skd}}
          \addplot[blue,mark=triangle*,line width=0.75pt] coordinates { (1,2.50)(4,2.478) (8,2.449) (12,2.432) (16,2.42) (20,2.413)};
          \addlegendentry{\textsc{Gpkd}}
          \end{axis}
      }
      \end{tikzpicture}
  }
  \hfill
  \subfigure[NIST Zh-En]
  {
      \centering
      \begin{tikzpicture}
      \scriptsize{
      \begin{axis}[
        ymajorgrids,
          xmajorgrids,
          grid style=dashed,
      width=.26\textwidth,
      height=.22\textwidth,
      legend style={at={(0.63,0.61)}, anchor=south west},
      xlabel={\scriptsize{Epoch}},
      ylabel={\scriptsize{Training Loss}},
      ylabel style={yshift=-2.4em},xlabel style={yshift=1.0em},
      yticklabel style={/pgf/number format/precision=1,/pgf/number format/fixed zerofill},
      ymin=2.8,ymax=4.2, ytick={3.0, 3.3, 3.6, 3.9},
      xmin=0,xmax=17,xtick={ 4, 8, 12, 16},
      legend style={at={(0.45,0.595)}, anchor=south west, legend plot pos=right,font=\tiny,cells={anchor=west}}
      ]

      \addplot[red,mark=otimes*,line width=0.75pt] coordinates {(3,4.10) (4,3.845) (6,3.467) (8,3.311) (10,3.22)(12,3.159)(14,3.116) (16,3.082)};
      \addlegendentry{\textsc{Skd}}
      \addplot[blue,mark=triangle*,line width=0.75pt] coordinates { (1,3.139) (4, 3.040) (8,2.976) (12,2.933) (16,2.904)};
      \addlegendentry{\textsc{Gpkd}}
      \end{axis}
      }
      \end{tikzpicture}
  }
  \caption{The training loss of applying the \textsc{Gpkd} (blue) and \textsc{Skd} (red) methods on the WMT En-De, NIST Zh-En tasks, respectively.}
  \label{fig:training-loss}
  \end{figure}

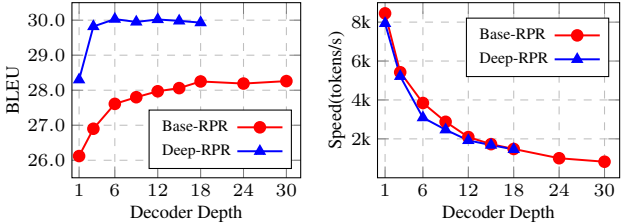
\begin{figure}[t]
  \small
  \subfigure
  {
      \centering
      \begin{tikzpicture}
      \scriptsize{
      \begin{axis}[
        ymajorgrids,
          xmajorgrids,
          grid style=dashed,
      width=.26\textwidth,
      height=.22\textwidth,
      legend style={at={(0.63,0.61)}, anchor=south west},
      xlabel={\scriptsize{Decoder Depth}},
      ylabel={\scriptsize{BLEU}},
      ylabel style={yshift=-1.8em},xlabel style={yshift=1.0em},
      yticklabel style={/pgf/number format/precision=1,/pgf/number format/fixed zerofill},
      ymin=25.5,ymax=30.5, ytick={26, 27, 28, 29, 30},
      xmin=0,xmax=32,xtick={1, 6, 12, 18, 24, 30},
      legend style={at={(0.335,0.05)}, anchor=south west, legend plot pos=right,font=\tiny,cells={anchor=west}}
      ]

      \addplot[red,mark=otimes*,line width=0.75pt] coordinates {(1,26.12) (3,26.90) (6,27.61)   (9,27.80) (12,27.97) (15, 28.06) (18,28.25) (24,28.19) (30,28.26)};
      \addlegendentry{Base-RPR}
      \addplot[blue,mark=triangle*,line width=0.75pt] coordinates { (1,28.30) (3,29.82) (6,30.03) (9,29.95) (12,30.02) (15,29.98) (18,29.93)};
      \addlegendentry{Deep-RPR}
      \end{axis}
      }
      \end{tikzpicture}
  }
  \hfill
  \subfigure
  {
      \begin{tikzpicture}
      \scriptsize{
          \begin{axis}[
            ymajorgrids,
          xmajorgrids,
          grid style=dashed,
          width=.27\textwidth,
          height=.22\textwidth,
          xlabel={\scriptsize{Decoder Depth}},
          ylabel={\scriptsize{Speed(tokens/s)}},
          ylabel style={yshift=-2.8em},xlabel style={yshift=1.0em},
          yticklabel style={/pgf/number format/precision=1,/pgf/number format/fixed zerofill},
          ymin=0,ymax=9, ytick={2, 4, 6, 8},
          yticklabel style={/pgf/number format/precision=0,/pgf/number format/fixed zerofill},
          yticklabel={$\pgfmathprintnumber{\tick}$k},
          xmin=0,xmax=32,xtick={1, 6, 12, 18, 24, 30},
          legend style={at={(0.365,0.575)}, anchor=south west, legend plot pos=right,font=\tiny,cells={anchor=west}}
          ]

          \addplot[red,mark=otimes*,line width=0.75pt] coordinates {(1,8.45026) (3, 5.424) (6,3.843) (9, 2.877) (12,2.0928) (15, 1.728)(18,1.48092) (24,1.00465) (30,0.82469)};
          \addlegendentry{Base-RPR}
          \addplot[blue,mark=triangle*,line width=0.75pt] coordinates {(1,7.935) (3,5.216) (6,3.092) (9,2.469) (12,1.921) (15,1.681) (18, 1.459)};
          \addlegendentry{Deep-RPR}
          \end{axis}
      }
      \end{tikzpicture}
  }
  \caption{BLEU scores [\%]  and translation speed  [tokens/sec] against decoder depth on the En-De task.}
  \label{fig:decoder-depth}
  \end{figure}

  \begin{table}[t]
    \small
    \setlength{\tabcolsep}{2.0pt}
    \centering
    \begin{tabular}{l c c c c c}
      \toprule
      \multicolumn{1}{l}{\multirow{2}{*}{\textbf{System}}}
      & \multicolumn{1}{c}{\multirow{2}{*}{\textbf{BLEU}}}
      & \multicolumn{2}{c}{\textbf{$\Delta$BLEU}}
      & \multicolumn{2}{c}{\textbf{Speedup}} \\
      \cmidrule(lr){3-4} \cmidrule(lr){5-6} 
      & & \textbf{Base} & \textbf{Deep} & \textbf{Base} & \textbf{Deep}\\
      \midrule
      \citet{zhang-etal-2018-accelerating}          &27.33   &-0.11  &N/A    &1.34$\times$    &N/A \\
      \citet{xiao2019sharing}                       &27.69   &+0.17  &N/A    &1.39$\times$    &N/A \\
      \midrule
      Base                                          &27.60   &N/A    &N/A    &N/A             &N/A\\
      Deep                                          &30.03   &N/A    &N/A    &N/A             &N/A\\
      \textsc{Desdar} 48-3                          &29.82   &+2.22  &-0.21  &1.28$\times$    &1.52$\times$ \\
      \textsc{Desdar} 48-1                          &28.31   &+0.71  &-1.72  &1.97$\times$    &2.17$\times$ \\
      \textsc{Desdar} 48-1 (\textsc{\textsc{Skd}})  &29.70   &+2.10  &-0.27  &1.99$\times$    &2.18$\times$ \\
      \textsc{Desdar} 48-1 (\textsc{Gpkd})          &30.01   &+2.41  &-0.02  &1.99$\times$    &2.18$\times$ \\
      \bottomrule
    \end{tabular}
    \caption{BLEU scores [\%], inference speedup [$\times$]  on the WMT En-De task.}
    \label{tab:speedup}
  \end{table}
  
For training, we used Adam optimizer \cite{kingma2014adam}, and followed the hyper-parameters uesd in \citet{wang-etal-2019-learning}. As suggested in \citet{shaw-etal-2018-self}, we incorporated the relative position representation into the self-attention mechanism to enhance the positional information. This is quite crucial when building extremely deep Transformer. Then, we batched sentence pairs by approximate length, and limited input/output tokens per batch to $4,096$/GPU and updated the parameters every two steps. The hidden size of Base and Deep models was 512, and 1024 for big counterparts. The Base/Big/Deep models were updated for 50k/150k/50k steps on the En-De task, 25k/50k/25k steps on the NIST Zh-En task and 100k/200k/100k on the WMT Zh-En task. The beam size and length penalty were set to 4/0.6 and 6/1.3 for En-De and Zh-En tasks, respectively. 


\subsection{Experimental Results}
\label{sec:gpkd}

\begin{table*}[!t]
  \small
  \centering
  \setlength{\tabcolsep}{2.2pt}
  \begin{tabular}{lrrrlcccccccc}
  \toprule

  \multirow{2}{*}{\textbf{System}} & \multicolumn{5}{c}{\textbf{WMT En-De}}  & \multicolumn{4}{c}{\textbf{NIST Zh-En}} & \multicolumn{3}{c}{\textbf{WMT Zh-En}} \\
  \cmidrule(r){2-6} \cmidrule(r){7-10}  \cmidrule(r){11-13}
  &{\bf Params}&{\bf Updates}&{\bf Time}&{\bf BLEU}&{\bf SBLEU}&{\bf MT06}&{\bf MT04} &{\bf MT05}&{\bf MT08} &{\bf Test17} &{\bf Test18}&{\bf Test19}\\

  \midrule
  \citet{fan2019reducing}        &286M  &16$\times$13K   &N/A       &30.20        &N/A       &N/A    &N/A         &N/A          &N/A        &N/A        &N/A        &N/A       \\
  \citet{li-etal-2020-shallow}   &194M  &50K   &11.75h       &30.21        &29.0       &N/A    &N/A         &N/A          &N/A        &N/A        &N/A        &N/A       \\
  \citet{wu-etal-2019-depth}     &270M  &800K   &N/A       &29.92        &N/A       &N/A    &N/A         &N/A          &N/A        &N/A        &N/A        &N/A       \\
  \citet{ott-EtAl:2018:WMT}      &210M  &16$\times$13K      &N/A       &29.30        &28.3      &N/A    &N/A         &N/A          &N/A        &N/A        &N/A        &N/A       \\
  \citet{wang-etal-2019-learning}&137M  &50K    &N/A       &29.30        &N/A       &53.57  &55.91       &52.30        &48.12      &26.9       &27.4       &N/A       \\
  \midrule
  Base-RPR                       &62M   &100K   &5.02h     &27.60        &26.5      &51.63  & 55.06      & 50.38       & 46.17     &25.9      &26.0      &27.6     \\
  Deep-RPR-48L                   &194M  &50K    &16.87h    &30.03        &28.8      &54.00  & 56.40      & 52.98       & 48.21     &26.8      &27.3      &28.6     \\
    + \textit{Skipping Sub-Layer}&194M  &50K    &15.41h    &\bf $\textrm{30.63}^{\diamond}$    &\bf 29.4  &\bf 54.76  &\bf 57.01   &\bf 53.52    &\bf 49.16  &\bf 27.2  &\bf 27.7  &\bf 29.0 \\
  \bottomrule
  \end{tabular}
  \caption{BLEU scores [\%], parameters and training time on three language pairs. Note that SBLEU represents the sacrebleu score and the $p_{l}$ (omission rate) was set to 0.4. Note that $\diamond$ presents the result of beam 8. The BLEU of beam 4 is 30.49.}\label{tab:main}
  \end{table*}

  \begin{figure*}
    \small
     \centering
     \subfigure[WMT En-De]
      {
      \begin{tikzpicture}
        \begin{axis}[
          ymajorgrids,
          xmajorgrids,
          grid style=dashed,
        width=.32\textwidth,
        height=.20\textwidth,
        legend style={at={(0.63,0.61)}, anchor=south west},
        xlabel={\scriptsize{Epoch}},
        ylabel={\scriptsize{Validation PPL}},
        ylabel style={yshift=-1.7em},xlabel style={yshift=0.7em},
        yticklabel style={/pgf/number format/precision=1,/pgf/number format/fixed zerofill},
        ymin=3.35,ymax=3.70, ytick={3.4, 3.5, 3.6},
        xmin=15.2,xmax=25.4,xtick={16, 18,  20,  22,  24},
        legend style={at={(0.2,0.68)}, anchor=south west,legend columns=2,legend plot pos=right,font=\tiny,cells={anchor=west}}
        ]
        \addplot[red,mark=otimes*,line width=0.75pt] coordinates {(16,3.52) (17,3.52) (18,3.54) (19,3.52) (20,3.53) (21,3.53) (22,3.55) (23,3.52) (24,3.56) (25,3.60)};
        \addlegendentry{Before}
        \addplot[blue,mark=triangle*,line width=0.75pt] coordinates {(16,3.48) (17,3.47) (18,3.45) (19,3.43) (20,3.42) (21,3.41) (22,3.40) (23,3.40) (24,3.39) (25,3.39)};
        \addlegendentry{After}
        \end{axis}
        \end{tikzpicture}
        }
     \centering
     \hspace{1em}
     \subfigure[NIST Zh-En]
      {
      \begin{tikzpicture}
        \begin{axis}[
          ymajorgrids,
          xmajorgrids,
          grid style=dashed,
        width=.32\textwidth,
        height=.20\textwidth,
        legend style={at={(0.90,0.61)}, anchor=south west},
        xlabel={\scriptsize{Epoch}},
        ylabel={\scriptsize{Validation PPL}},
        ylabel style={yshift=-1.7em},xlabel style={yshift=0.7em},
        yticklabel style={/pgf/number format/precision=1,/pgf/number format/fixed zerofill},
        ymin=4.1,ymax=4.6, ytick={4.2, 4.3, 4.4, 4.5},
        xmin=8.8,xmax=16.2,xtick={10, 11, 12, 13, 14, 15},
        legend style={at={(0.2,0.68)}, anchor=south west, legend columns=2,legend plot pos=right,font=\tiny,cells={anchor=west}}
        ]
        \addplot[red,mark=otimes*,line width=0.75pt] coordinates {(9,4.52) (10,4.43)  (11,4.35)  (12,4.36)  (13,4.32)  (14,4.35)  (15,4.37)  (16,4.38) };
        \addlegendentry{Before}
        \addplot[blue,mark=triangle*,line width=0.75pt] coordinates {(9,4.37) (10,4.25)  (11,4.23)  (12,4.22)  (13,4.19)  (14,4.18)  (15,4.17)  (16,4.17) };
        \addlegendentry{After}
        \end{axis}
        \end{tikzpicture}
      }
      \centering
      \hspace{1em}
      \subfigure[WMT Zh-En]
      {
      \begin{tikzpicture}
          \begin{axis}[
            ymajorgrids,
          xmajorgrids,
          grid style=dashed,
            width=.32\textwidth,
          height=.20\textwidth,
          legend style={at={(0.63,0.61)}, anchor=south west},
          xlabel={\scriptsize{Epoch}},
          ylabel={\scriptsize{Validation PPL}},
          ylabel style={yshift=-1.7em},xlabel style={yshift=0.7em},
          yticklabel style={/pgf/number format/precision=1,/pgf/number format/fixed zerofill},
          ymin=4.85,ymax=5.20, ytick={4.9, 5.1},
          xmin=9.8,xmax=16.2,xtick={10, 11, 12, 13, 14, 15},
          legend style={at={(0.2,0.68)}, anchor=south west,legend columns=2, legend plot pos=right,font=\tiny,cells={anchor=west}}
          ]

          \addplot[red,mark=otimes*,line width=0.75pt] coordinates {(10,5.08)  (11,5.07)  (12,5.03)  (13,5.01)  (14,5.02)  (15,5.04)  (16,5.06) };
          \addlegendentry{Before}
        \addplot[blue,mark=triangle*,line width=0.75pt] coordinates {(10,5.01)  (11,4.99)  (12,4.94)  (13,4.92)  (14,4.91)  (15,4.91)  (16,4.90) };
        \addlegendentry{After}
          \end{axis}
      \end{tikzpicture}
      }
    \caption{The validation PPL of employing the Skipping Sub-Layer method before (red) and after (blue) on En-De, NIST Zh-En and WMT Zh-En tasks, respectively.}
    \label{fig:valid ppl}

  \end{figure*}
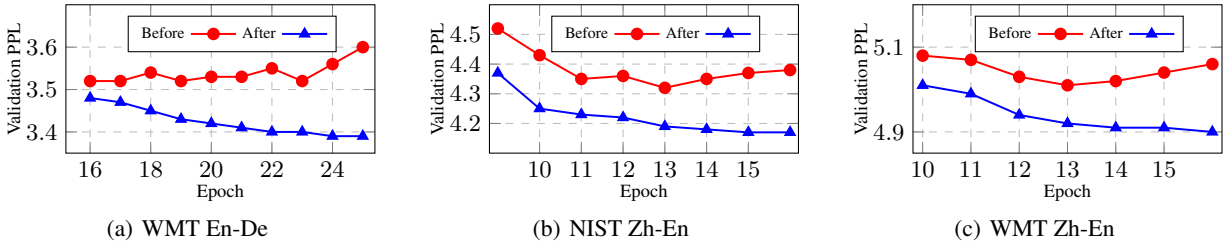

\subsubsection{The Effect of \textsc{Gpkd} Method}
We successfully trained 24-layer/48-layer Transformer-Deep systems and 12-layer Transformer-Big systems incorporating the relative position representation (RPR) on three tasks. Table \ref{tab:compression} shows the results when applying the \textsc{Gpkd} method to the encoder side. Deep Transformer systems outperform the shallow baselines by a large margin, but the model capacities are 2 or 3 times larger. And 6-layer models trained through \textsc{Skd} outperform the shallow baselines by $0.63$-$1.39$ BLEU scores, but there is still a nonnegligible gap between them and their deep teachers. As we expect, our \textsc{Gpkd} method can enable the baselines to perform similarly with the deep teacher systems, and outperforms \textsc{Skd} by $0.41$-$1.10$ BLEU scores on three benchmarks. Note that, although the compressed systems are 4 or 8 $\times$ shallower, they only underperform the deep baselines by a small margin. Similar phenomenon is observed when switching to a wide network, that 6-layer RPR-Big systems match with its teacher with almost no loss in BLEU, indicating the \textsc{Gpkd} method is applicable in different model capacities.  Moreover, Figure \ref{fig:training-loss} plots the training loss of \textsc{Skd} and \textsc{Gpkd} methods. We observe that \textsc{Gpkd} obtains a much lower training loss than \textsc{Skd} on the WMT En-De and NIST Zh-En tasks, which further verifies the effectiveness of \textsc{Gpkd}.


Figure \ref{fig:decoder-depth} plots the BLEU scores and the translation speeds of different decoder depths. We can see that deeper decoders yield modest BLEU improvements but slow down the inference significantly based on a shallow encoder. While no improvement is observed when we switch to a strong Deep-RPR-48L model. 

Table \ref{tab:speedup} exhibits several \textsc{Desdar} systems with different settings. \textsc{Desdar} 48-3 achieves comparable performance with the 48-6 baseline, but speeds up the inference by $1.52\times$. However, a shallower decoder makes a great decrease compact on BLEU, though it obtains a $1.97\times$ speedup. Through the \textsc{Skd} method, the \textsc{Desdar} 48-1 system can even outperform the RPR-Base by $2.10$ BLEU scores and speeds up the inference by $2.18\times$. Moreover, our \textsc{Gpkd} method can enable the \textsc{Desdar} 48-1 system to perform similarly with the deep baseline, outperforms \textsc{Skd} by nearly +$0.31$ BLEU scores. Interestingly, after knowledge distillation, the beam search seems like to be not important for the \textsc{Desdar} systems, which can achieve a $3.2\times$ speedup with no performance sacrifice with the greedy search. This may be due to the fact that the student network learns the soft distribution generated by the teacher network, which has already limited the search space to the max beam margin \cite{kim-rush-2016-sequence}.

\subsubsection{The Effect of Skipping Sub-Layer Method}
 \label{sec:model-comparison}


The red curves in Figure \ref{fig:valid ppl} show that the 48-layer RPR model converges quickly on three tasks, and the validation PPL goes up later. At the same time, the training PPL is still going down (see Figure \ref{fig:train-valid ppl}). As we expect, the \textit{Skipping Sub-Layer} method reduces the overfitting problem and thus achieves a lower PPL (3.39) on the validation set. The similar phenomena are observed on the other two tasks. In addition, the last row of Table \ref{tab:main} shows that the strong Deep-RPR model trained through the \textit{Skipping Sub-Layer} approach obtains +$0.40$-$0.72$ BLEU improvements on three benchmarks. 

\subsubsection{Comparison with Related Methods}

Table \ref{tab:comparison} exhibits the BLEU scores and the validation PPL of several related systems trained through two optimization ways. Interestingly, all these systems underperform the deep baseline when we trained them from scratch. This is reasonable because the skipped connections make disturbances to the optimization when the model is still in the early stage of training, and the parameters are in the non-smoothed region of the loss function. The phenomena here verify the importance of the two-stage training strategy. 

On the other hand, our \textit{Skipping Sub-Layer} method and \textit{Stochastic Layers} \cite{pham2019very} trained by the finetuning schema both beat the strong baseline. This confirms that dropping sub-layers randomly is helpful for reducing the overfitting when we train deep Transformer models. However, the deep models trained with \textit{LayerDrop} method cannot gain more benefit and we attribute this to the fact that \textit{LayerDrop} uses the same probability to omit sub-layers throughout the stack. This is harmful to the performance because omitting many lower-level layers reduces the representation ability of the deep model significantly \cite{huang2016stochastic,iclrGreff2017}.

\subsubsection{Ablation Study}
 
Table \ref{tab:ablation} shows the ablation study of omitting different components, including randomly skipping the feed-forward (FFN) sub-layer, the self-attention (SAN) sub-layer, all sub-layers and the whole layer. As shown in Table \ref{tab:ablation}, the performance of the single checkpoint and the checkpoint averaging model is reported. First, we can see that all these systems obtain lower validation PPLs and higher BLEU scores for the single model than the baseline. And our default strategy beats the baseline by a larger margin in terms of both PPL and BLEU. There is no significant difference when we skip FFN or SAN only, and they all surpass the system that randomly omits the entire layer. This is mainly due to the fact that we can sample more diverse sub-networks in training. Note that the results here were mainly experimented on deep Transformer, rather than a shallow but wide counterpart reported in \cite{fan2019reducing}, which is complementary to the community.

\begin{table}[t]
  \small
  \setlength{\tabcolsep}{3pt}
  \centering
  \begin{tabular}{rl|rr}
  \toprule
  \multicolumn{2}{l|}{\textbf{System}}
  & \bf PPL&\bf BLEU\\
  \midrule
  \multicolumn{2}{l|}{Deep-RPR-48L}  &3.52       &30.03\\
  \cmidrule(){1-4}
  \multirow{3}{*}{\rotatebox{90}{Scratch}}
  & \multicolumn{1}{|l|}{\textit{Skipping Sub-Layer}}                        &3.41       &29.82\\
  & \multicolumn{1}{|l|}{\textit{LayerDrop} \cite{fan2019reducing}}          &3.42       &29.65\\
  & \multicolumn{1}{|l|}{\textit{Stochastic Layers} \cite{pham2019very}}     &3.44       &29.75\\
  \cmidrule(){1-4}
  \multirow{3}{*}{\rotatebox{90}{Finetune}}
  & \multicolumn{1}{|l|}{\textit{Skipping Sub-Layer}}                        &\bf 3.39       &\bf 30.49\\
  & \multicolumn{1}{|l|}{\textit{LayerDrop} \cite{fan2019reducing}}          &3.47       &29.77\\
  & \multicolumn{1}{|l|}{\textit{Stochastic Layers} \cite{pham2019very}}     &3.44       &30.12\\
  \bottomrule
  \end{tabular}
  \caption{Comparison of training from scratch and finetune on the WMT En-De task. }\label{tab:comparison}
\end{table}

\begin{table}[t]
  \small
  \setlength{\tabcolsep}{10.0pt}
  \centering
  \begin{tabular}{l|rcc}
  \toprule
  \textbf{System} & \textbf{PPL}  & \textbf{Single} & \textbf{Avg} \\
  \midrule
  Deep-RPR-48L      & 3.52        &29.19        &30.03 \\ 
  \midrule    
  Skip FFN          & 3.45        &29.61        &30.22\\
  Skip SAN          & 3.45        &29.71        &30.26\\
  Skip FFN or SAN   & \bf 3.39    &\bf 29.95    &\bf30.49\\
  Skip Layer        & 3.46        &29.45        &29.92\\
  \bottomrule
  \end{tabular}
  \caption{Ablation results on the WMT En-De task.}
  \label{tab:ablation}
  \end{table}

\begin{table}[t]
  \small
  \setlength{\tabcolsep}{3pt}
  \centering
  \begin{tabular}{l r r r r}
    \toprule
    \textbf{System} & \textbf{Params} & \textbf{Depth} & \textbf{BLEU} &\textbf{Speed} \\
    \midrule
    \citet{fan2019reducing}             &286M               &12-6      &30.20    &2534 \\
    \citet{wu-etal-2019-depth}          &270M               &8-8       &29.92    &2044 \\
    \citet{wei2004multiscale}            &512M               &18-6      &30.56    &N/A  \\
    \midrule
    \textit{Skipping Sub-Layer} + \textsc{Gp}        &194M               &48-6      &30.59    &3092 \\
    \textsc{Gpkd} 24-3                           &106M               &24-3      &30.40    &5237 \\
    \textsc{Gpkd} 24-1                           &97M                &24-1      &30.05    &8116 \\
    \textsc{Gpkd} 6-6                            &62M                &6-6       &30.16    &3817 \\
    \textsc{Gpkd} 6-3                            &48M                &6-3       &29.71    &5460 \\

    \bottomrule
  \end{tabular}
  \caption{The overall results of BLEU scores [\%] and translation speed [tokens/sec] on the WMT En-De task.}
  \label{tab:overall}
\end{table}

\subsubsection{The Overall Results}

Table \ref{tab:overall} shows the results of incorporating both the \textsc{Gpkd} and \textit{Skipping Sub-Layer} approaches. Note that, these systems are obtained upon the strong Deep-RPR-48L system. As we can see that a 6-6 system achieves comparable performance with the state-of-the-art, though the parameter is only 4 times less than theirs. In addition, it beats the shallow baseline by +$2.56$ BLEU scores at the same scale. This offers a way of selecting the proper system considering the trade-off between the translation performance and the model storage. For example, one can choose \textsc{Gpkd} 6-3 system with satisfactory performance and fast inference speed, or \textsc{Gpkd} 24-3 system with both high translation quality and competitive inference speed. Another interesting finding here is that shrinking the decoder depth may hurt the BLEU score when the encoder is not strong enough.

\section{Related Work}
\label{sec:related work}

Deep neural networks play an important role in the resurgence of deep learning. It has been observed that increasing the depth of neural networks can drastically improve the performance of convolutional neural network-based systems \cite{he2016deep}. The machine translation communities follow this trend. For example, \citet{bapna-etal-2018-training} and \citet{wang-etal-2019-learning} shortened the path from upper-level layers to lower-level layers so as to avoid gradient vanishing/exploding. \citet{wu-etal-2019-depth} designed a two-stage approach with three specially designed components to build a 8-layer Transformer-Big system. \citet{zhang-etal-2019-improving} successfully trained a deep Post-Norm Transformer with carefully designed layer-wise initialization strategy. More attempts on initialization strategy emereged recently \cite{xu2019lipschitz,liu2020understanding,Huang2020improving}. Perhaps the most relevant work with us is \citet{fan2019reducing}'s work. They employed \textit{LayerDrop} mechanism to train a 12-6 Transformer-Big and pruned sub-networks during inference without finetuning. Here we address a similar issue in deep Transformer, which has not been discussed yet. Beyond this, we present a new training strategy that can boost the deep system in a robust manner.

For model compression, there are many successful methods, such as quantization \cite{gong2014compressing}, knowledge distillation (\textsc{Kd}) \cite{kim-rush-2016-sequence}, weight pruning \cite{Han2015learning} and efficient Transformer architecture \cite{mehta2020delight,mehta2020define}. For Transformer models, \citet{sun-etal-2019-patient} proposed a novel approach to compressing a large BERT model into a shallow one via the Patient Knowledge Distillation method. \citet{jiao2019tinybert} achieved a better compression rate by richer supervision signals between the teacher network and the student network. However, these methods are not straightforwardly applicable to machine translation, they need simultaneously compute the logits of each layer in both the teacher and student networks, which consumes large GPU memory. In this work, we propose the \textsc{Gpkd} method to compress an extremely deep model into a baseline-like system, without any additional computation cost.

\section{Conclusions}
\label{sec:conclusions}

Our contributions in this work are two folds. (\romannumeral1) We propose a \textsc{Gpkd} method to compress the deep model into a shallower one with minor performance sacrifice, which outperforms the \textsc{Skd} method by a large margin. (\romannumeral2) The proposed \textit{Skipping Sub-Layer} method reduces the overfitting problem when training extremely deep encoder systems by randomly omitting sub-layers during training phase. The experimental results on three widely-used benchmarks validate the effectiveness of the proposed methods. After the incorporating of two methods, the strong but light-weight student models show competitive performance which is application friendly.

\section*{Acknowledgments}
This work was supported in part by the National Science Foundation of China (Nos. 61876035 and 61732005), the National Key R\&D Program of China (No. 2019QY1801). The authors would like to thank anonymous reviewers for their valuable comments. And thank Qiang Wang and Yufan Jiang for their helpful advice to improve the paper. Appreciate Tao Zhou for his suggestion to beautify the figures.

\bibliography{GPKD}
\end{document}